\newcommand{\ARXIV}{} 

\ifdefined\ARXIV
\documentclass[letterpaper, 10pt, twocolumn]{article}
\usepackage[margin=0.75in]{geometry}
\usepackage{libertine}
\renewcommand\footnotemark{} 
\usepackage{hyperref}
\hypersetup{colorlinks = true, citecolor = {black}, linkcolor = black}
\date{}
\else

\documentclass[letterpaper, conference]{IEEEtran}
\IEEEoverridecommandlockouts
\let\labelindent\relax 
\fi

\usepackage{algorithm} 
\usepackage{algorithmicx} 
\usepackage{algpseudocode}

\usepackage[pdftex]{graphicx}
\usepackage{amssymb}
\usepackage{amsmath}
\usepackage{cleveref} 
\usepackage{subfigure}
\usepackage{balance}
\usepackage[export]{adjustbox}

\title{\LARGE \bf
Impedance Adaptation by Reinforcement Learning \\ with Contact Dynamic Movement Primitives
}

\author{Chunyang Chang$^{1, \dagger}$, Kevin Haninger$^{2, \dagger}$, Yunlei Shi$^{4, 1}$, Chengjie Yuan$^{3}$, Zhaopeng Chen$^{1}$, Jianwei Zhang$^{4}$
\thanks{$^{1}$ with Agile Robots AG, Munich, Germany. Email:
        {\tt\small chunyang.chang@agile-robots.com}}%
\thanks{$^{2}$ with the Department of Automation at Fraunhofer IPK, Berlin, Germany}%
\thanks{$^{3}$ with Intel Asia-Pacific Research \& Development Ltd., Shanghai, China}
\thanks{$^{4}$ with TAMS (Technical Aspects of Multimodal Systems), Department of Informatics, Universit\"at Hamburg, Hamburg, Germany}
\thanks{$^{\dagger}$ The first two authors contributed equally to this work. This research has received funding from the German Research Foundation (DFG) and the National Science Foundation of China (NSFC) in project Crossmodal Learning, DFG TRR-169/NSFC 61621136008, partially supported by the European Union’s Horizon 2020 research and innovation programme under the Marie Sklodowska-Curie grant agreement No 778602 ULTRACEPT and No 820689 SHERLOCK.}%
}

\begin{document}

\maketitle
\thispagestyle{empty}
\pagestyle{empty}


\begin{abstract}

Dynamic movement primitives (DMPs) allow complex position trajectories to be efficiently demonstrated to a robot. In contact-rich tasks, where position trajectories alone may not be safe or robust over variation in contact geometry, DMPs have been extended to include force trajectories. However, different task phases or degrees of freedom may require the tracking of either position or force -- e.g., once contact is made, it may be more important to track the force demonstration trajectory in the contact direction. The robot impedance balances between following a position or force reference trajectory, where a high stiffness tracks position and a low stiffness tracks force. This paper proposes using DMPs to learn position and force trajectories from demonstrations, then adapting the impedance parameters online with a higher-level control policy trained by reinforcement learning. This allows one-shot demonstration of the task with DMPs, and improved robustness and performance from the impedance adaptation.  The approach is validated on peg-in-hole and adhesive strip application tasks. 

\end{abstract}


\section{INTRODUCTION}
A major challenge in robotics is reducing the commissioning effort for a new task. For free-space tasks, the task is mostly solved by a position trajectory of the robot, which can be taught by a variety of methods \cite{ravichandar2020}. However, contact-rich tasks, which require the application of force on a possibly uncertain environment, are typically not solved with only position control.  

Some tasks, e.g. polishing tasks, are more robustly represented as force tasks, requiring a force trajectory  to be applied in certain degrees of freedom (DOF).  Other tasks require a mix of position and force control, which can be considered in a unified way with impedance control \cite{hogan1985impedance}.  In impedance control, a high impedance (i.e. high stiffness) better tracks a position trajectory, and a low impedance better tracks a force trajectory \cite{bicchi2004}.  

Dynamic movement primitives (DMPs) allow the efficient teaching of a trajectory - only one demonstration is needed. In the original formulation \cite{ijspeert2002}, DMPs track a position trajectory, allowing a trajectory shape to be generalized over changes in initial or goal positions. DMPs have also been extended into contact DMPs \cite{nemec2013}, where force trajectories are also recorded. This allows the DMP to be employed with an impedance- or force-controlled robot \cite{abu-dakka2015, kober2015}, improving safety and robustness over small changes in geometry. 

However, different impedance behaviors are required in different applications -- due to, e.g. different part geometries, contact materials, or tolerances. Impedance may also vary during a task as contact conditions change \cite{abu-dakka2020}. Adaptation of impedance is also important for humans, especially in unstable tasks like screwdriving \cite{li2018}.  

\begin{figure}
    \centering
    \includegraphics[width=0.8\columnwidth]{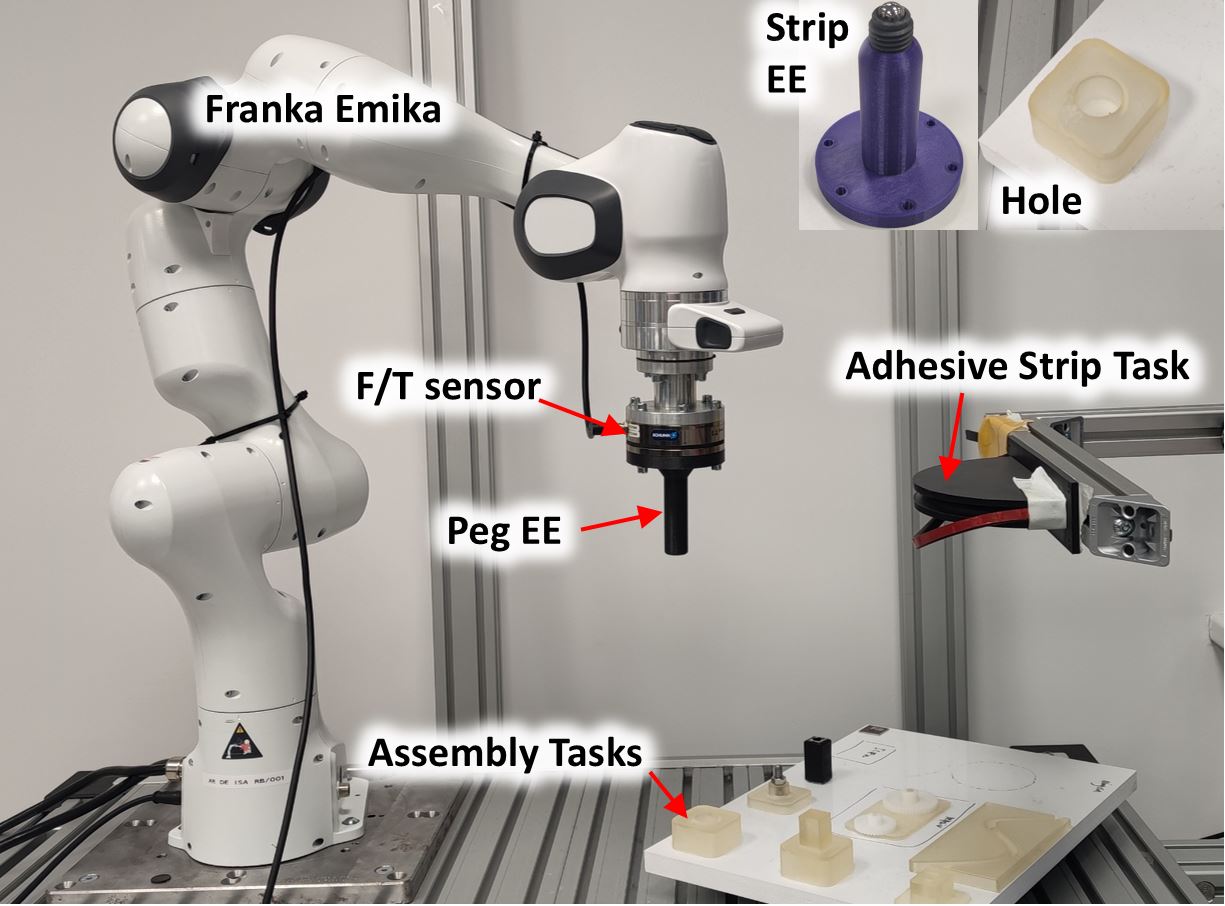}
    \caption{Experimental setup, including robot, additional F/T sensor, peg-in-hole and adhesive tape assembly task. Insets on top right show the hole and the adhesive strip end-effector. \label{fig:Experiment_environment}}
\end{figure}

Optimizing impedance parameters can improve the cycle time and robustness of a task \cite{beltran-hernandez2020}, reduce force overshoot \cite{roveda2018} or reduce trajectory jerk \cite{dimeas2015}. There are a wide variety of application-specific and ad-hoc methods for tuning impedance, recently reviewed in \cite{abu-dakka2020}. Determining robot impedance from demonstrations can be done by transferring from human demonstrations \cite{tang2015, abu-dakka2018}, an additional human interface \cite{peternel2018robotic}, determined by a model predictive controller \cite{haninger2021a}, or determined by variance of the demonstrations \cite{calinon2010}.  

Reinforcement learning (RL) also aims to reduce manual commissioning work, allowing the robot to explore and learn a control policy. To improve sample efficiency -- how much trial-and-error is required -- RL has been combined with DMPs. In \cite{buchli2011, stulp2012}, a DMP trajectory is taught and a position control gain schedule is learned. RL has also been applied to learn a force trajectory \cite{hazara2016} or a residual force trajectory \cite{davchev2022} for contact-rich tasks.

Compared with other contact DMPs \cite{abu-dakka2015}, we learn online adaptation of the impedance to balance the tracking of force and position trajectories. Compared with other combinations of DMPs and RL, this paper uses RL to adapt impedance, not learn a force trajectory \cite{hazara2016, davchev2022}. This paper first introduces the teaching workflow and system architecture, then presents the individual modules (admittance control, dynamic movement primitives, and reinforcement learning), then presents the experimental results.

\section{Problem Statement and System Architecture}
\begin{figure}
    \centering
    \includegraphics[width=\columnwidth]{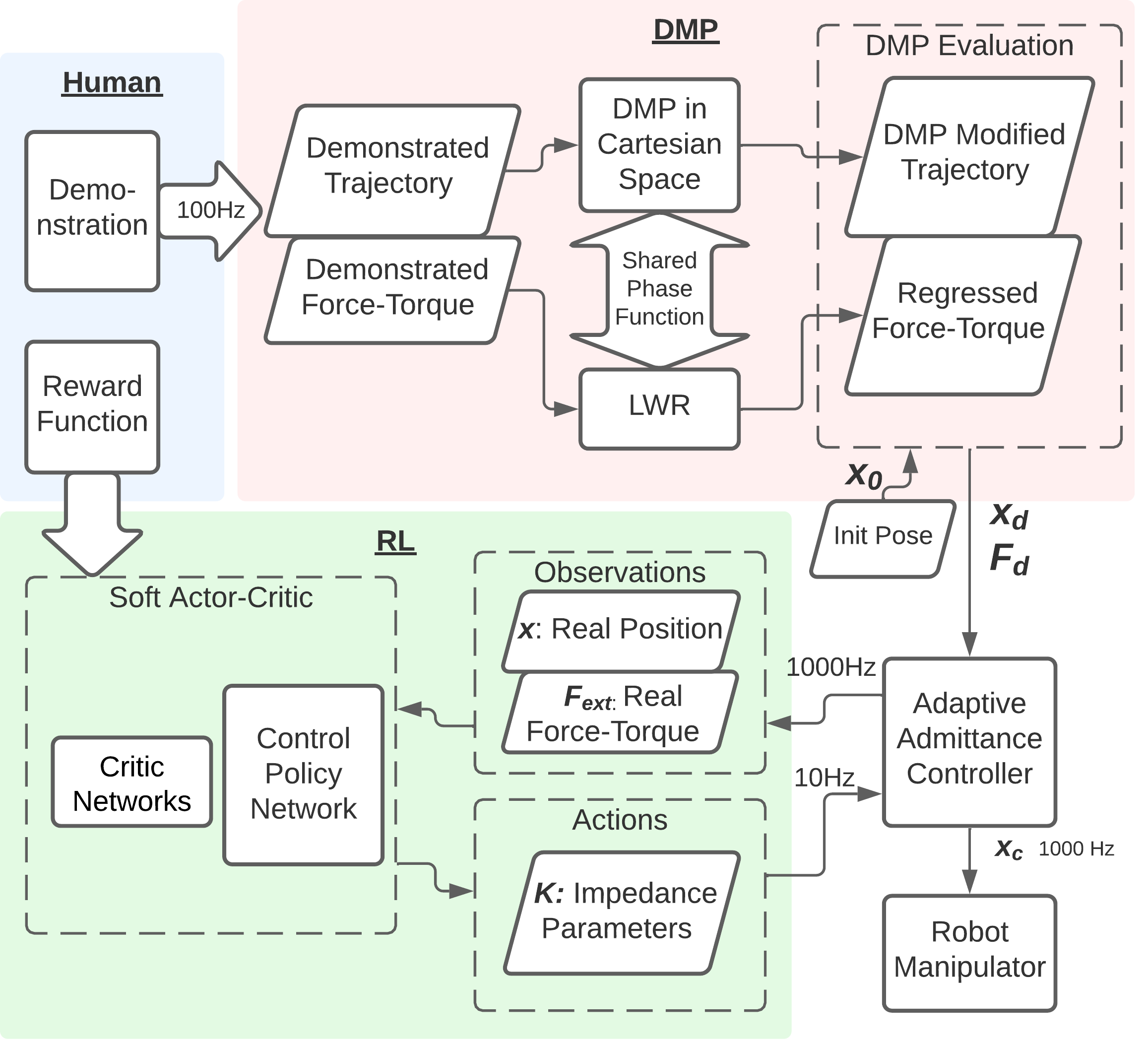}
    \caption{Data flow diagram for the complete system}
    \label{fig:Architecture_of_Pipeline}
\end{figure}

A robot with a hand guiding mode and a force/torque (F/T) sensor, as seen in Figure \ref{fig:Experiment_environment} is used.  The commissioning process here contains three stages:
\begin{enumerate}
    \item {\bf Demonstration Collection:} Using the hand guidance mode of the robot, the robot is grasped on the robot side of the F/T sensor and task is demonstrated. The position data from the robot is acquired along with the measurements from the F/T sensor.
    \item {\bf Autonomous Data Collection:} The demonstration is processed into a DMP, which is executed autonomously.  A control policy which maps position and force to a change in robot impedance is learned by the soft actor-critic RL algorithm.
    \item {\bf Execution:} The system is executed, with the DMP producing the position and force trajectory, and the RL policy adjusting impedance according to measured position and force.
\end{enumerate}
An overview of data flow during the commissioning process can be seen in Figure \ref{fig:Architecture_of_Pipeline}.  The individual modules (admittance controller, DMPs, and RL policy), will be introduced in the following sections.

\section{Admittance Control}
To allow data collection where human and environment forces are separated, a F/T sensor is used, as labelled in Figure \ref{fig:Experiment_environment}. The human forces for the hand guidance in demonstration are measured by the built-in joint torque sensors of the robot. The environmental contact forces are measured with the flange F/T sensor during the demonstration. This separation of human and environment forces is established for contact-rich demonstrations \cite{tang2015, hazara2016}. 

\subsection{Admittance Controller}
\begin{figure}
    \centering
    \includegraphics[width=0.8\columnwidth]{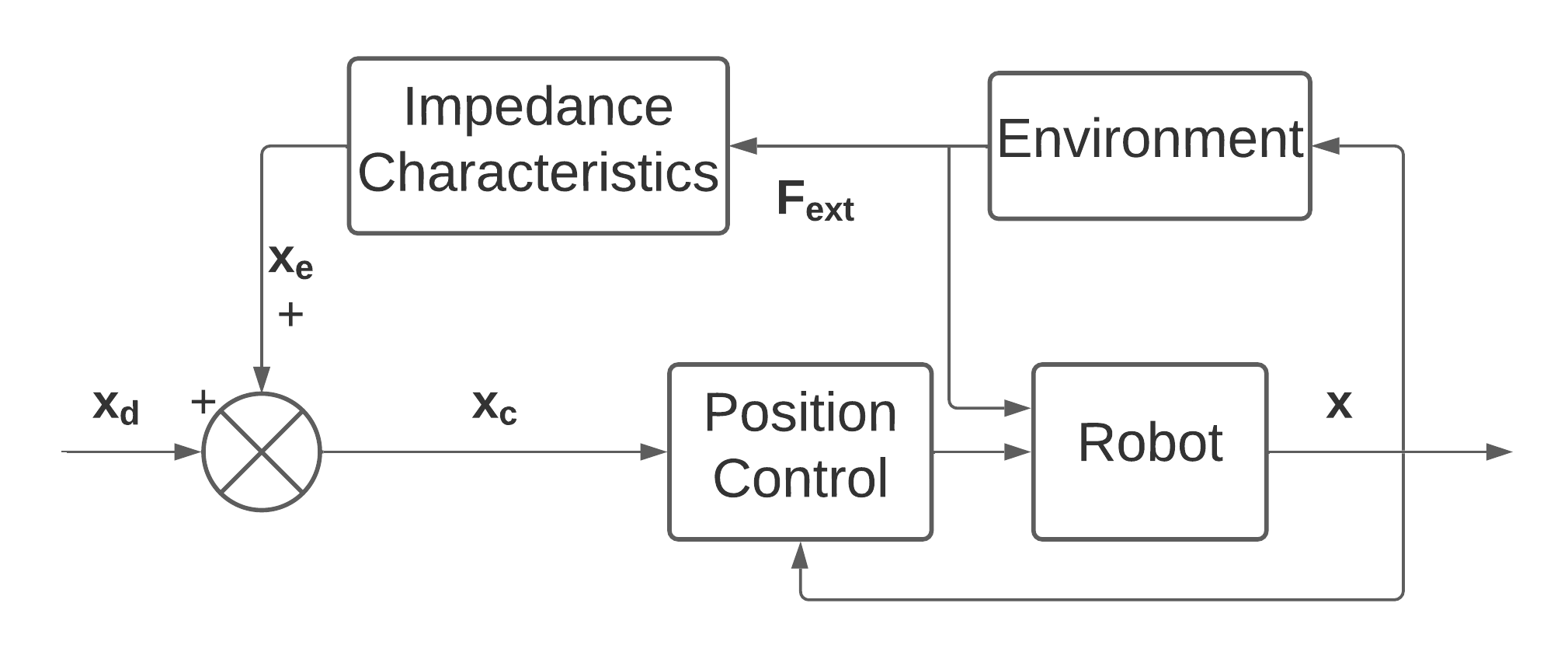}
    \caption{Control block diagram of the admittance controller}
    \label{fig:Admittance_controller}
\end{figure}

The admittance control law of 
\begin{equation}\label{equation:adm_computation_1}
    \mathbf{ M}\ddot{\mathbf{x}}_e+\mathbf{D}\dot{\mathbf{x}}_e+\mathbf{K}\mathbf{x}_e = \mathbf{F}_{ext}-\mathbf{F}_d
\end{equation}
is realized, where $\mathbf{x}_e$ is the pose deviation from planned trajectory $\mathbf{x}_d$, $\mathbf{F}_{ext}$ is the measured external force and $\mathbf{F}_d$ is the desired force-torque, all in the TCP coordinate frame. The matrices $\mathbf{M}$, $\mathbf{D}$ and $\mathbf{K}$ are control parameters, diagonal $6\times 6$ matrices. The admittance controller can be seen in Figure \ref{fig:Admittance_controller}.

\subsubsection{Trade-off between position and force control}
To motivate the use of impedance stiffness $\mathbf{K}$ to modulate between force and position control, we examine \eqref{equation:adm_computation_1} for a single DOF and various $K$. Taking the Laplace domain form of \eqref{equation:adm_computation_1}, we find $Ms^2+Ds+K = F_{ext}(s)-F_d(s)$.  To find the steady-state behavior, we consider $s\rightarrow 0$, and constant $F_{ext}$, $F_d$.

When $K\rightarrow \infty$, $x_e = K^{-1}(F_{ext}-F_d)$, and it can be seen that small $x_e$ results, so the commanded robot position is almost equal the position trajectory, i.e. $x_c \approx x_d$. When $K\rightarrow 0$, $F_{ext}-F_d \approx 0$, so the real force $F_{ext}$ tracks the desired force $F_d$.  

\subsubsection{Orientation implementation}
The linear and orientation components of a pose $\mathbf{ x}_{[\cdot]}=[\mathbf{x}^p_{[\cdot]}, \mathbf{q}_{[\cdot]}]$, where $\mathbf{q}_{[\cdot]}$ is a unit quaternion. A unit quaternion is $\mathbf{q} = v + \mathbf{u} \in \mathcal{S}^3$, where $\mathcal{S}^3$ is a unit sphere in $\mathbb{R}^4$, consisting of a scalar part $v \in \mathbb{R}$ and a vector part $\mathbf{u} \in \mathbb{R}^3$. The multiplication, conjugation and norm of quaternions are defined as
\begin{align}
        \mathbf{q}_1 * \mathbf{q}_2 &= (v_1 v_2 \texttt{-} \mathbf{u}^\top_1 \mathbf{u}_2) \texttt{+} (v_1\mathbf{u}_2 \texttt{+} v_2 \mathbf{u}_1 \texttt{+} \mathbf{u}_1 \times \mathbf{u}_2) \label{equation:quaternion_multi}\\ 
        \overline{\mathbf{q}} &= \overline{v + \mathbf{u}} = v - \mathbf{u} \label{equation:quaternion_conjugation} \\
      ||{\mathbf{q}}|| &= \sqrt{\mathbf{q} * \overline{\mathbf{q}}}= \sqrt{v^2 + ||\mathbf{u}||^2} ,      \label{equation:quaternion_norm}
\end{align}
where $||\mathbf{u}||$ is the $\ell_2$ norm of $\mathbf{u}$. To map $\mathcal{S}^3$ to $\mathbb{R}^3$, the quaternion logarithm function $\log : \mathcal{S}^3 \mapsto \mathbb{R}^3$ is given by
\begin{equation}\label{equation:log_mapping}
    \log(\mathbf{q}) = \log(v + \mathbf{u}) = 
        \begin{cases}
        \arccos(v)\frac{\mathbf{u}}{||\mathbf{u}||}, & ||\mathbf{u}|| \neq 0\\
        [0, 0, 0]^\top, & otherwise
        \end{cases}
\end{equation}
Furthermore, there is a relationship between the quaternion logarithm and the angular velocity as
\begin{equation}
    \label{equation:error_angular_velocity}
    \boldsymbol{\omega} = 2\Delta_t^{-1} \log(\mathbf{q}_{t + 1} * \overline{\mathbf{q}_{t}}),
\end{equation}
where $\boldsymbol{\omega}\in\mathbb{R}^3$ denotes the angular velocity that rotates quaternion $\mathbf{q}_{t}$ into $\mathbf{q}_{t + 1}$ within time $\Delta_t$.

Inversely, the exponential map $\exp : \mathbb{R}^3 \mapsto \mathcal{S}^3$ is
\begin{equation}
    \label{equation:exp_mapping}
    \exp(\mathbf{r}) = 
    \begin{cases}
        \cos(||\mathbf{r}||) + \sin(||\mathbf{r}||)\frac{\mathbf{r}}{||\mathbf{r}||}, & ||\mathbf{r}||\neq0\\
        1+[0, 0, 0]^\top, &otherwise
    \end{cases}
\end{equation}

With the inverse of \eqref{equation:error_angular_velocity}, the new orientation $\mathbf{q}_{t + 1}$ can also be derived with the known orientation $\mathbf{q}_t$ and angular velocity $\boldsymbol{\omega}$ by
\begin{equation}
    \label{equation:control_calculate_orientation}
    \mathbf{q}_{t + 1} = \exp(\frac{\boldsymbol{\omega}\Delta_t}{2})\mathbf{q}_{t}
\end{equation}

The orientation of admittance controller in \eqref{equation:adm_computation_1} is calculated by solving for angular acceleration $\dot{\mathbf{\omega}}_{e,t}$, integrating to angular velocity $\mathbf{\omega}_{e,t}$, then using \eqref{equation:control_calculate_orientation} to update the orientation $\mathbf{q}_{e,t}$ to $\mathbf{q}_{e,t+1}$. 

Finally, the commanded pose $\mathbf{x}_c$ could be specified by the multiplication of desired pose and error pose, which is:
\begin{equation}\label{computation_4}
    \mathbf{X}_c = \mathbf{X}_d \mathbf{X}_e
\end{equation}
where $\mathbf{X}_{[\cdot]}$ is the $4\times4$ transformation matrix of corresponding $\mathbf{x}_{[\cdot]}$. The order is switched because $\mathbf{X}_e$ is in TCP coordinates, and $\mathbf{X}_d$ is in base coordinates.

\subsection{Admittance Gains}
The impedance characteristics in \eqref{equation:adm_computation_1} are determined by the parameters $\mathbf{M}$, $\mathbf{D}$ and $\mathbf{K}$ with the equation. This can be rewritten as
\begin{equation}
\mathbf{D} = 2\zeta\sqrt{\mathbf{MK}} \label{eq:damping_equation}
\end{equation}
where $\zeta$ is a damping parameter, $\zeta > 1$ is overdamping and $\zeta < 1$ underdamping with more oscillation. 

Here, the $M^p$ (mass) are $5$ kg for the linear components and the orientation components $M^o$ (inertia) are $0.02$  kgm$^2$. The value of $\mathbf{K}$ is assigned by the RL policy, and is thus allowed to be updated online. The critical damping is chosen by \eqref{eq:damping_equation} with $\zeta=1$.

To maintain contact stability, limits on linear stiffness $K^p \in [20, 2000]$ and orientation stiffness $K^o \in [1, 40]$ were used.  A maximum change in the admittance controller of $|K^p_{t+1}-K^p_t| < 40$ and $|K^o_{t+1}-K^o_t| < 1$ were enforced.

\section{Dynamic Movement Primitives with Force}
This section explains the construction and evaluation of the DMPs, including both rotation and force information.  
\subsection{Position Dynamic Movement Primitives}
A DMP is a representation for complex motor actions, motivated by differential equations of well-defined attractor dynamics \cite{schaal2006dynamic}. It is defined by point attractor dynamics
\begin{equation}\label{equation:dmp}
    \tau^2\ddot{x} = \alpha_x(\beta_x(g - x) - \tau\dot{x})  + f,
\end{equation}
where $x$ is the position of the system, $g$ is the goal position, $\dot{x}$ and $\ddot{x}$ are velocity and acceleration, $\alpha_x$ and $\beta_x$ are gain terms that determine the damping and spring behavior of the system, $f$ is a forcing function, and $\tau$ adjusts the temporal behavior of the system.

The forcing function $f$ that models the nonlinear behavior is learned as a function of phase variable $z$. It can be formulated using a linear combination of $N$ nonlinear Radial Basis Functions (RBFs) as 
\begin{equation}\label{equation:formulate_fz}
    f(z) = (g - x_0)\sum_{i=1}^N \psi_i(z) w_i z,
\end{equation}
where $x_0$ is the initial pose of the robot, and the term $(g - x_0)$ is to maintain the shape of the trajectory when the goal $g$ is changed, and
\begin{equation}\label{equation:psi_z}
\psi_i(z) = \frac{\exp{(-h_i(z - c_i)^2)}}{\sum_{j = 1}^N\exp{(-h_j(z - c_j)^2)}},
\end{equation}
where $h_i$ and $c_i$ are constants that determine the width and centers of the basis functions, respectively, and are manual parameters. Setting $c_i= z_0\exp{(-\frac{\alpha_z}{\tau}\frac{i\cdot T}{N})}$ makes an even distribution of RBFs over time, and the basis function bandwidth used here is $h_i=\frac{N^{1.5}}{c_i \cdot \alpha_z}$, found by trial and error.

The phase variable $z$ evolves as
\begin{equation}\label{equation:calculate_z}
    z = z_0\exp{(-\frac{\alpha_z}{\tau}t)},
\end{equation}
where $\alpha_z$ is a constant controlling the speed, $t$ is the time step in $\{0, 1, 2, \dots, T\}$, and $z_0$ is the initial value of $z$.

The target value of the forcing function is calculated from the demonstration as 
\begin{equation}\label{equation:calculate_ftarget}
    f_{tar}(t) = \tau^2\ddot{x}_{demo}(t) \texttt{-} \alpha_x(\beta_x(g \texttt{-} x_{demo}(t)) \texttt{-} \tau\dot{x}_{demo}(t)).
\end{equation}
Thus, in \eqref{equation:formulate_fz}, the weights $\mathbf{w} = [w_1, w_2, w_3, \dots, w_N]^T$ over basis functions could be learned to make the forcing term match $f_{tar}(t)$ by a regression algorithm, such as locally weighted regression (LWR) \cite{cleveland1988locally}, giving a fit weight vector $\mathbf{w}$ of
\begin{equation}\label{equation:calculate_w}
    \mathbf{w} = (\mathbf{\Psi}^\top\mathbf{\Psi})^{-1}\mathbf{\Psi}^\top\mathbf{F}.
\end{equation}
For the attractor dynamics, $\mathbf{\Psi}=[\psi_1,\dots,\psi_T]$ and $\mathbf{F}$ is given by
\begin{equation}\label{equation:calculate_F}
    \mathbf{F} = [\frac{f_{tar}(0)}{(g - x_0)z(0)},\dots, \frac{f_{tar}(T)}{(g - x_0)z(T)}]^\top.
\end{equation}

\subsection{Orientation with Quaternion Representation}

The algorithms mentioned above are designed for the single DOF motion. To handle rotations, the quaternion form of DMPs is used \cite{ude2014orientation}. Thus, the expression of \eqref{equation:dmp} in quaternion form is given by
\begin{equation}\label{equation:quaternion_dmp}
    \tau^2\dot{\boldsymbol{\omega}} = \alpha_x(\beta_x 2\log(\mathbf{g}^o * \overline{\mathbf{q}}) - \boldsymbol{\omega}) + f^o(z),
\end{equation}
where $\boldsymbol{\omega}$ is the angular velocity and $\textbf{g}^o$ is the quaternion representation of goal orientation. The corresponding forcing function also has a similar form as \eqref{equation:formulate_fz}
\begin{equation}
    \label{equation:formulate_fz_orientation}
    f^o(z) = \mathrm{diag}(\log(\mathbf{g}^o * \overline{\mathbf{q}}))\sum_{i=1}^N \psi_i(z) w_i^o z,
\end{equation}
with orientation weights $w_i^o$, which are fit in the same way as \eqref{equation:calculate_ftarget} and \eqref{equation:calculate_w}.

\subsection{Representation of Force-Torque Information}
To integrate the force-torque demonstration information, linear combinations of RBFs are utilized to regress the forces over Cartesian space as \cite{abu-dakka2015}
\begin{equation}
    \label{equation:formulate_fz_F}
    F_d(z) = \sum_{i=1}^N \psi_i(z) w_i^F
\end{equation}
where the $z$ and $\psi_i(z)$ are calculated by the \eqref{equation:calculate_z} and \eqref{equation:psi_z}, and $w_i^F$ are weights for the force/torque regression, calculated as in \eqref{equation:calculate_w}.

\subsection{DMP Commissioning and Execution}
To commission, i.e. to find weights $\mathbf{w}^p,\mathbf{w}^o,\mathbf{w}^F$, first, the phase variable $z_t$ is calculated with \eqref{equation:calculate_z}, the RBFs $\psi$ of the forcing function $f$ are calculated using \eqref{equation:psi_z}, velocity and acceleration calculated by the numerical derivative and \eqref{equation:error_angular_velocity}, the forcing functions are calculated with \eqref{equation:calculate_ftarget}, then the weight vector $\mathbf{w}$ is found by \eqref{equation:calculate_w}. The parameters used here are $\alpha_z=1$, $\tau=25$, $\alpha_y=6.25$.

To execute, the phase variable $z$ is calculated by \eqref{equation:calculate_z}, the forcing functions are calculated \eqref{equation:formulate_fz} and \eqref{equation:formulate_fz_orientation}, then the position and orientation trajectory by numerical integration on \eqref{equation:dmp} and \eqref{equation:quaternion_dmp}.  The desired force is calculated by \eqref{equation:formulate_fz_F}.

\section{Reinforcement Learning}
\label{sec:Reinforcement_learning}
The soft actor-critic (SAC) algorithm optimizes a stochastic policy in an off-policy way \cite{haarnoja2018b}. The significant characteristic of SAC is determining exploration noise, how much the actions deviate from the currently optimal action, by entropy regularization. At a high level, SAC empirically finds an approximately optimal policy $\pi^*:S\times A\rightarrow [0,1]$, a distribution over state $\mathbf{s}\in S$ and action $\mathbf{a} \in A$, as  
\begin{equation}
    \pi^* = \arg\max_\pi \sum_{t=0}^T \mathbb{E}_{\mathbf{s}_t,\mathbf{a}_t} r(\mathbf{s}_t,\mathbf{a}_t) + \alpha\mathcal{H}(\pi(\cdot|\mathbf{s}_t)),
\end{equation}
where $\mathcal{H}(\pi(\cdot | \mathbf{s}_t)) = \int \pi(\mathbf{a}_t|\mathbf{s}_t)\log(\pi(\mathbf{a}_t|\mathbf{s}_t)d\mathbf{a}_t$ is a measure for how random the policy $\pi$ is, and coefficient $\alpha >0$ increase the exploration noise. 

SAC is here used to adapt the impedance according to the current force and position. The state and actions for the agent are
\begin{align}
\mathbf{s} &= [\mathbf{x}^p, \mathbf{q}, \mathbf{F}_{ext}] \\
\mathbf{a} &= [\mathbf{K}^p, \mathbf{K}^o],
\end{align} 
where $\mathbf{x}^p \in \mathbb{R}^3$ is linear position, $\mathbf{q} \in \mathcal{S}^3$ is the orientation quaternion, $\mathbf{F}_{ext} \in \mathbb{R}^6$ is the force/torque measurement, and $\mathbf{K}^p, \mathbf{K}^o \in \mathbb{R}^{3}$ are the linear and orientation stiffness parameters, where $\mathbf{K}=\mathrm{diag}([\mathbf{K}^p, \mathbf{K}^o])$. 

The SAC policy and critic networks are three linear fully-connected layers, with 256 nodes per layer and ReLU activation functions. An entropy term of $\alpha=0.05$ is used with learning rate scheduled from $3.3e-3$ to $3e-4$. Other parameters are standard from \cite{raffin2021stable}.

Inspired by \cite{beltran2020variable}, the reward function is designed as 
\begin{equation}
\begin{split}
    \label{equation: reward function}
        r(\mathbf{s}, \mathbf{a})  = & \, \omega_{goal}\kappa(\mathbf{s})+\omega_{p}L(||\mathbf{x}^p-\mathbf{x}^p_d||/e_{term,p}) \\
         & + \omega_qL(e^o/e_{term,q}) + \omega_FL(||\mathbf{F}^p_{ext}-\mathbf{F}_d^p||/e_{term,F}) \\
         &  +  \omega_ML(||\mathbf{F}^o_{ext}-\mathbf{F}^o_d||/e_{term,M}) 
\end{split}
\end{equation}
where $F^p_{[\cdot]}$ are linear forces, $F^o_{[\cdot]}$ are the torques, $e_{term,\cdot}$ is the defined allowed maximal difference, $L(x) = 1 - x/\sqrt{3}$ to map to the range $0$ to $1$, $\omega_{[\cdot]}$ are weighting hyperparameters, and $e^o$ is the orientation error in logarithm form as
\begin{equation}
    \label{equation: distance of orieantation}
    e^o = 
    \begin{cases}
    2 \pi, &\mathbf{q} * \overline{\mathbf{q}}_d = -1 + [0, 0, 0]^\top\\
    2||\log(\mathbf{q} * \overline{\mathbf{q}}_d)||_2, &otherwise 
    \end{cases}
\end{equation}
because the logarithmic map has no discontinuity boundary except for a singularity at $\mathbf{q}*\overline{\mathbf{q}}_{d} = -1 + [0, 0, 0]^\top$. 

Weighting hyperparameters of $\omega_{goal}=1$ and $\omega_{p}=\omega_{q}=\omega_{F}=\omega_{M}=0.5$ are used here.  The terminating distance $e_{term,p}=0.01$, $e_{term,q}=0.1$, $e_{term,F}=3.0$ and $e_{term,M}=1.0$.
The first term $\kappa(\mathbf{s})$ is the completion term, which is defined as 
\begin{equation}
    \label{equation: kappa}
    \kappa(\mathbf{s}) = 
    \begin{cases}
        100,    &task \ finished\\
        -50,    &terminated\\
        -100,   &error\\
        0,      &otherwise
    \end{cases}
\end{equation}

The purpose of this term is to encourage the agent to complete the task. If the difference of any element in full state $[\mathbf{x}^p, \mathbf{q}, \mathbf{F}_{ext}]$ exceeds the corresponding $e_{term,\cdot}$, the movement would be terminated and the status would be labeled as $terminated$; if an error such as joint discontinuity occurs and robot is stopped during moving, the status is labeled as $error$; if the movement is eventually completed, the status is then $finished$; otherwise, there is no external reward added. 

\section{Experimental Validation}
This section presents the implementation and evaluation on two contact-rich tasks.
 \subsection{Implementation}
 \subsubsection{Hardware}
For the robot, Franka Emika Panda robot has been used with the Cartesian pose interface at 1kHz. An external force-torque sensor SCHUNK Axia80 has been mounted on the flange of the robot arm and connected to the higher-level computer with an Ethernet connection. Then two different 3d printed end effectors are mounted on the other side of the sensor according to the experiments.

\subsubsection{Software}
To transfer data between sensor, robot and computer, the middleware software Links and Nodes (LN, developed by DLR) is used. LN is similar to ROS, and provides two communication paradigms, publishing and service calls. For controlling the Franka robot, the Franka Control Interface (FCI) with C++ APIs has been used.

To store the raw and modified data, Sqlite is used as the SQL database engine. The reason is that it is flexible and provides Python and C++ APIs so that it is possible to operate the data in both programming languages.

We used Gym \cite{brockman2016openai} as the framework of RL with a custom environment and Stable Baseline 3 \cite{raffin2021stable} as the implementation of RL algorithm SAC. 
 
\subsection{Peg in hole task}
We used insertion task seen in Figure \ref{fig:Experiment_environment} to test and evaluate the performance of the full state representation of position and force trajectories. The diameters of the peg and the hole are 20mm and 20.5mm, respectively, and $K^p=600$, $K^o=13$ for the tests.

\subsubsection{DMP in Cartesian space}
\label{subsubsection:DMP in Cartesian space}
\begin{figure}
    \centering
    \includegraphics[width=\columnwidth]{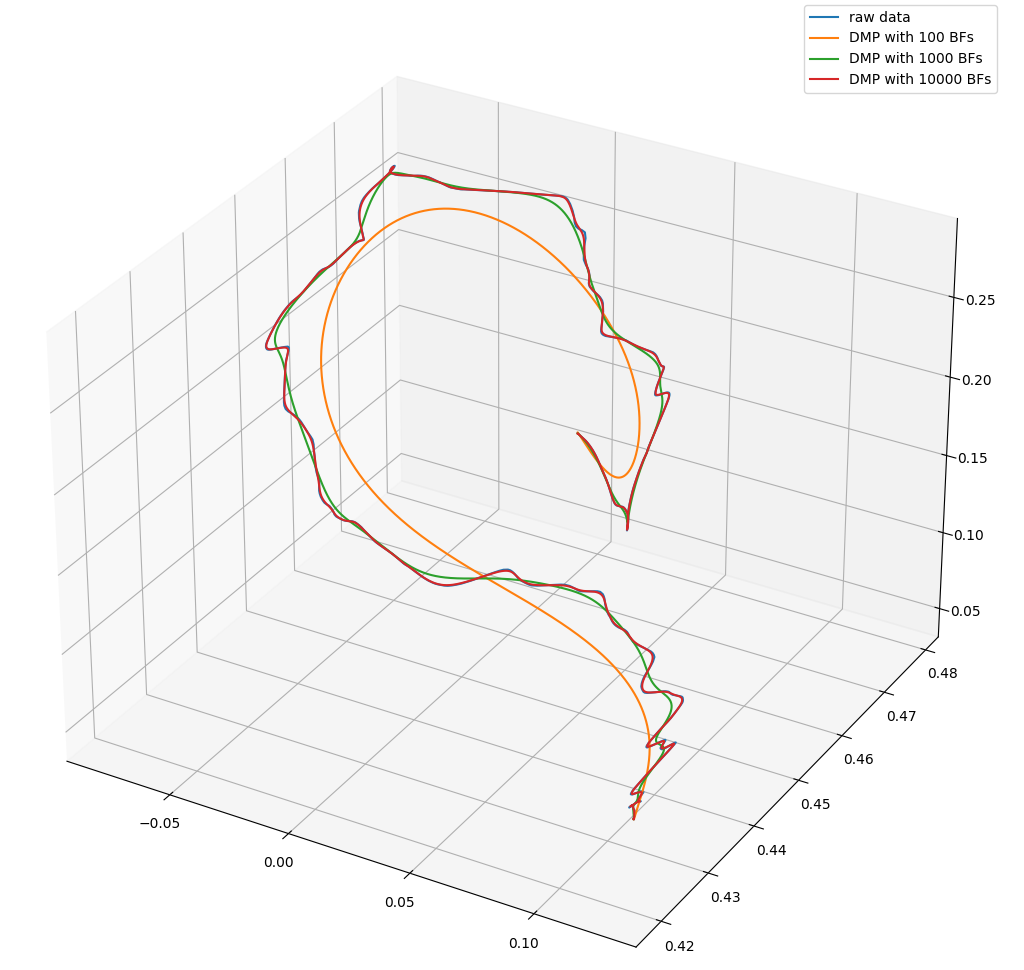}
    \caption{Translational trajectories comparison in 3d view, axes units are meters}
    \label{fig:3d_plot}
\end{figure}

Firstly, the DMP algorithm in Cartesian free space has been tested. After a demonstration without force-torque involved, the raw demonstrated trajectory has been recorded. Then the DMP modified trajectories have been generated according to different number of basis functions. To better visualize the results, we represent the translation of each trajectory in the 3d view as follows in Figure \ref{fig:3d_plot}.

As shown in Figure \ref{fig:3d_plot}, the more basis functions there are, the higher the similarity will be. In this case the trajectory with 10,000 BFs shows barely any difference compared to raw data. If the number of BFs is relatively smaller such as 1,000, the corresponding trajectory becomes smoother and maintains the characteristics of the original trajectory to a certain extent. With a much smaller number of BFs (100), the trajectory is the smoothest but is significantly different from the raw trajectory. 

\subsubsection{Force-Torque Representation}
\begin{figure}
    \centering
    \includegraphics[width=0.7\columnwidth]{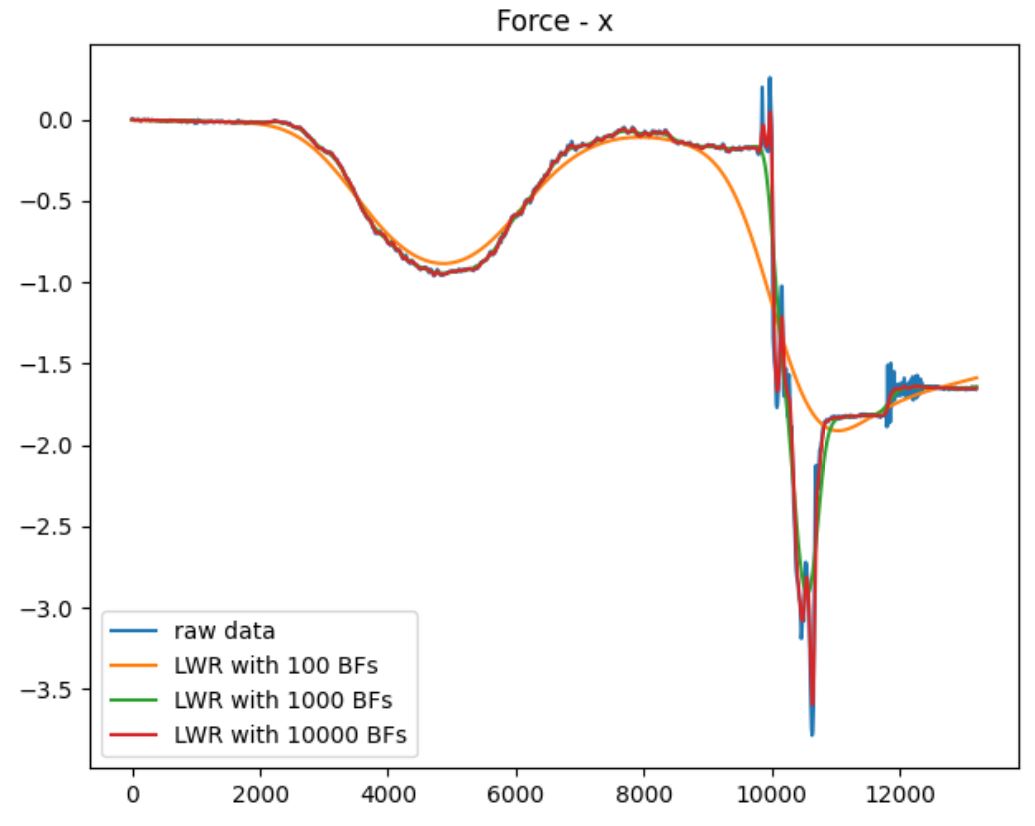}
    \caption{Force-Torque Profiles in X-axis, Raw Data and DMP Data with Different Number of Basis Functions. The x-axis is milliseconds, and the y-axis is $N$.}
    \label{fig:ft_plot}
\end{figure}

We investigate if the LWR force regression is suited to hard contact. To test this, we then demonstrate a trajectory with contact, and compare LWR generated force-torque profiles with a range of number of basis functions, seen in Figure \ref{fig:ft_plot}. 

The force curve changes quickly due to the environment being stiff, where typical overshoots from contact transitions can be seen in the raw data (blue). The 10,000 BFs most closely matches the raw data, but can cause errors when used on the robot (joint acceleration discontinuity error).  The 100 BF shows the highest smoothness but lacks accuracy.

\subsubsection{Force and Position Regression}
To test the force-torque part explicitly, we pushed the end-effector down a bit more and then move back after inserting the hole to get a proper value of force, and performed insertion tests using 100, 1,000 and 10,000 BFs. 100 trials have been proceeded for each setting to obtain the success rate. Success is defined by a final z error of $<5$ mm (hole depth is $10$ mm). The results can be seen in Table \ref{tab:Exp_results}.

\begin{table*}
    \caption{Success rate in different experimental conditions}
    \centering
    \begin{tabular}{|r|l|r|l|r|l|}
        \hline
        \multicolumn{2}{|c|}{\bf Peg-in-Hole}  & \multicolumn{2}{|c|}{\bf Tape, original} &    \multicolumn{2}{|c|}{\bf Tape, with 1 mm shift} \\
        \hline
        \# BFs   & Succ. &  Stiff.    ($K^p$, $K^o$)       & Succ.    & Stiff.  ($K^p$, $K^o$)         & Succ.  \\
        \hline
        100             & 92\%         &  low(50, 0.5)        &  0\%            & low(50, 0.5)        &  0\%          \\
        \hline
        1,000            & 79\%         &  middle(605, 13)     & 72\%            & middle(605, 13)     & 61\%          \\
        \hline
        10,000           & 50\%         & high(2000, 40)       & 96\%            & high(2000, 40)      & 73\%          \\
        \hline
             -           &  -            & SAC                  & 97\%            & SAC                 & 93\%          \\
        \hline
    \end{tabular}
    \label{tab:Exp_results}
\end{table*}

\begin{figure}[htbp]
    \centering
    \includegraphics[width=\columnwidth]{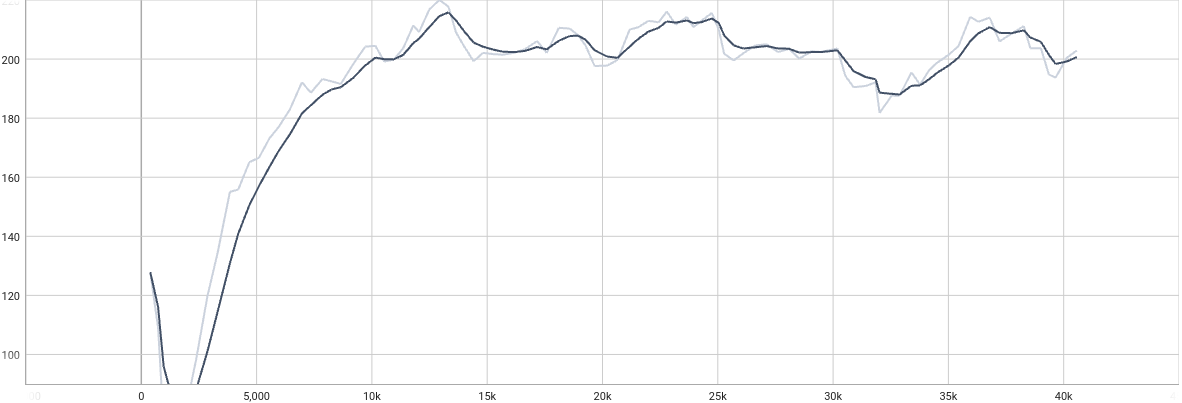}
    \caption{Learning Curve of SAC, with average $\sum r$ on the y-axis and training steps on x-axis.}
    \label{fig:learning_curve}
\end{figure}

From Table \ref{tab:Exp_results}, we see that the approach can achieve a good success rate, but modifying the number of BFs affects the success rate. More BFs can cause stability issues during contact due to more frequent changes in force (as seen in Figure \ref{fig:ft_plot}). The pose tracking is not as sensitive to the number of BFs as the pose varies more slowly over time. 

\subsection{Adhesive strip application task}
\label{section:tape task}

A task of pressing a tape into a press fit along a long, curved channel, as shown in Figure \ref{fig:Experiment_environment}, using the tape end effector in the inset of Figure \ref{fig:Experiment_environment} which has a smooth steel ball at the end. To allow the experiment to reset automatically, both ends of the tape are fixed to the fixture.

\subsubsection{Evaluation of SAC Algorithm}

First, the raw demonstration data were converted to the DMP with 500 BFs. The SAC is applied with parameters seen in Section \ref{sec:Reinforcement_learning}, and the learning curve can be seen in Figure \ref{fig:learning_curve}, where $40k$ training steps takes about $2$ hours, including reset time.

Then we evaluated the trained RL model compared with the trajectories with different fixed stiffness and get the success rate by 100 trials, with results shown in the middle column of Table \ref{tab:Exp_results}. The fixed stiffness values are given for $K^p$ and $K^o$. 

The success rate depends strongly on stiffness, where generally the success rate increases with stiffness as this prevents the end-effector from slipping off the tape. For the SAC agent, stiffness varies over time as can be seen in Figure \ref{fig:Tape_task_Kpy}, which shows how the stiffness in the $y$ direction varies during the contact-rich task. The $y$ direction is along the tape. 

We can analyze the difference between actual force and desired force in y axis, shown in Figure \ref{fig:Tape_task_diff}. From this figure, the most risky time period for Fy in this task is around 10s to 11s, because $F_d$ increases rapidly in this time, sometimes causing a discontinuity error. During this time period, the stiffness increases, and the force trajectory is followed less, so the risk of discontinuity error is reduced.

The stiffness in the three Cartesian directions are plotted in Figure \ref{fig:Tape_task_Kp_compare}, where the $x$ is across the tape, $y$ along the tape, and $z$ into the tape. During the main contact motion of the task, $K^p$ is higher in $x$, keeping from slipping off the tape. $K^p$ is lower in $z$, which is the contact direction, doing more force control.  Near 10 seconds, the gains are affected by the discontinuity error and breaking contact with the tape.

\subsubsection{Evaluation of Robustness}
The ability to tolerate small position errors in assembly is a key feature. To test this, we manually added an offset of 1 mm to the base coordinate system $y+$ axis in the controller, making the end-effector closer to the fixture.  The obtained results are shown in Table \ref{tab:Exp_results}.

\begin{figure}
    \centering
    \subfigure[$K^p$ in y\label{fig:Tape_task_Kpy}]{\includegraphics[width=\columnwidth]{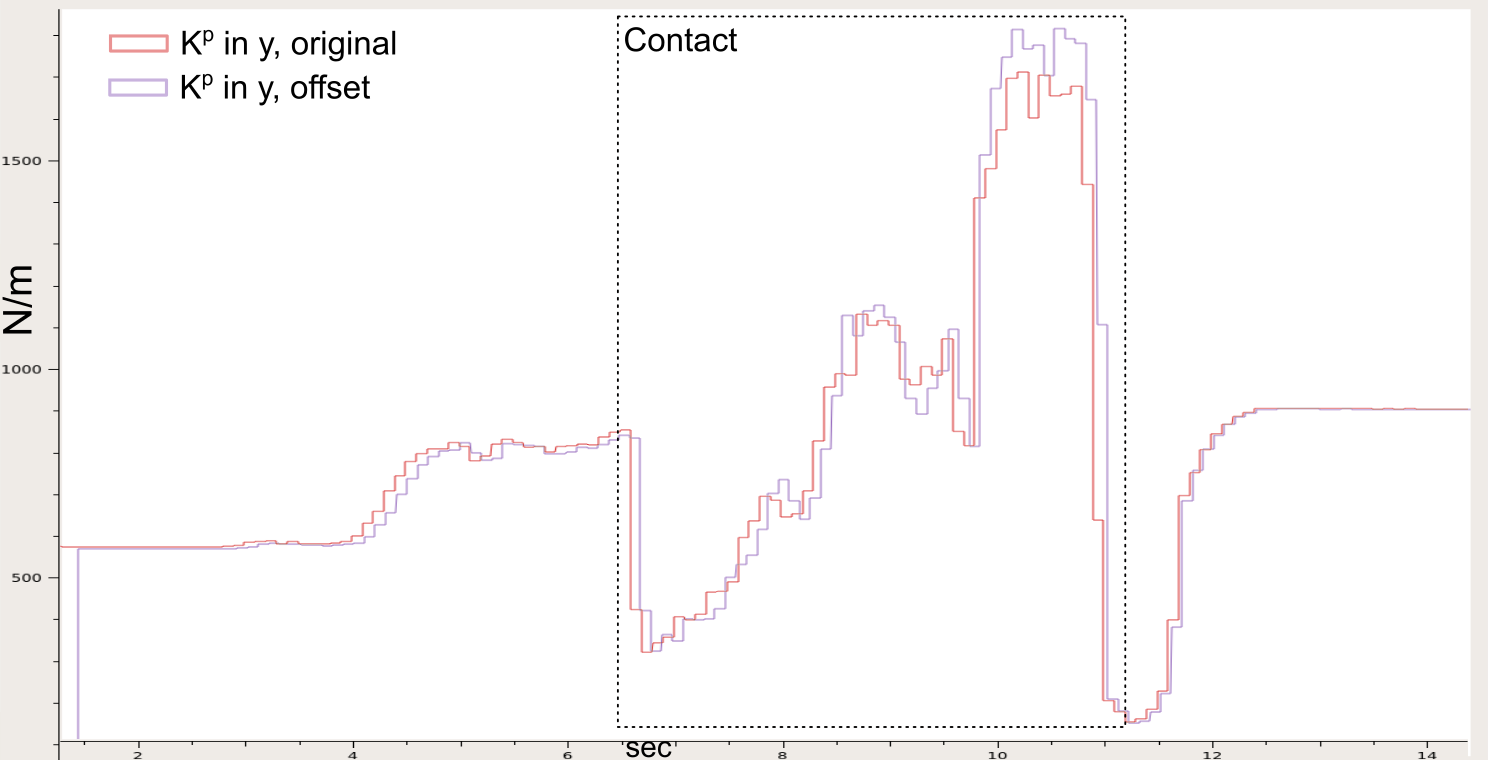}} \\
    \subfigure[$F_{ext}-F_d$ in y\label{fig:Tape_task_diff}]{\includegraphics[width=\columnwidth]{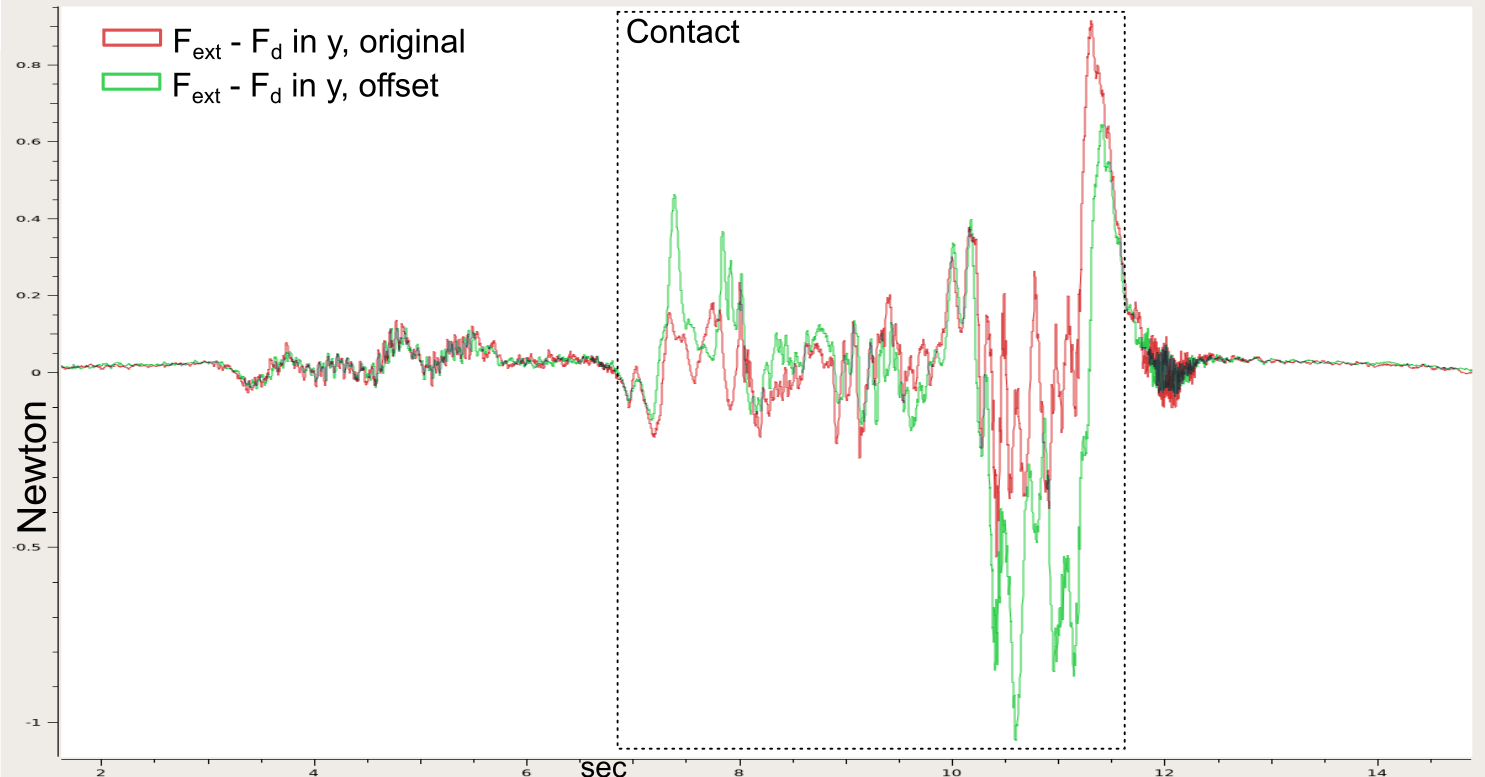}} \\
    \subfigure[$K^p$ in x, y, and z\label{fig:Tape_task_Kp_compare}]{\includegraphics[width=\columnwidth]{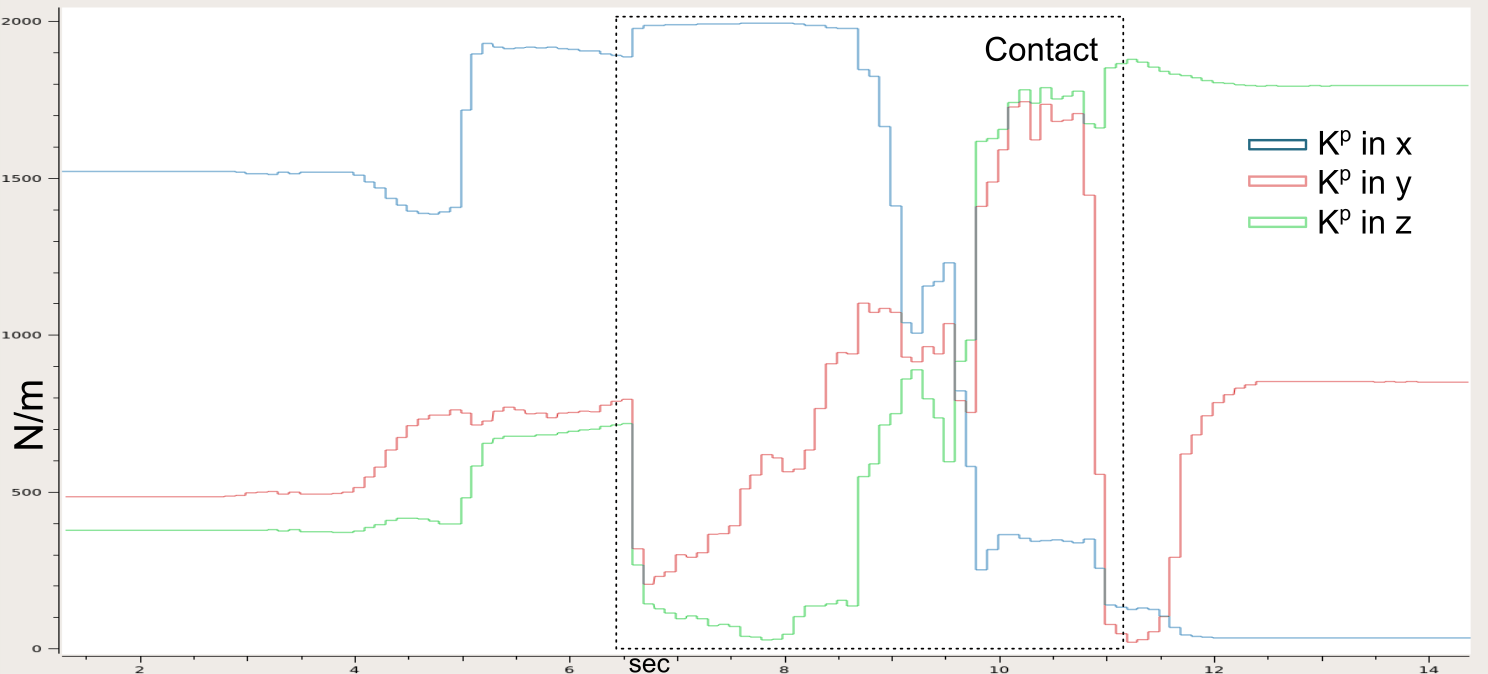}}
    \caption{Trained SAC agent on tape task, where (a) shows $K^p$ on original and offset, (b) shows the force error $F_{ext}-F_d$ on original and offset, and (c) shows the Cartesian $K^p$ on original.}
    \label{fig:Tape_task_SAC_kp2}
\end{figure}

After adding the offset, on the one hand, the success rate of all experimental groups has decreased because of the bigger contact force and higher probability of impacting the fixture; on the other hand, the success rate with SAC model has not decreased very much compared with other groups. 

To explain the generalization of the model more clearly, we plotted $\mathbf{F}_{ext}-\mathbf{F}_d$ and stiffness in the $y$ direction in the offset scenario and compared it with the original scenario in Figures \ref{fig:Tape_task_Kpy} and \ref{fig:Tape_task_diff}. These plots are manually time aligned, and show that a similar profile is reached even though the timing was originally offset.  Slightly larger forces near 10-12 seconds in the offset task cause higher stiffness to ensure successful completion of the task.

\section{CONCLUSION}
In this work, we proved that the proposed framework with the combination of admittance controller, DMPs and RL is feasible as an efficient way to teach contact-rich tasks. On the one hand, the DMPs could smooth the demonstrated position and force trajectories, meanwhile provide an up-sampling function to generate the trajectories from lower to higher frequency. On the other hand, the RL policy could balance the ratio of position and force control by adapting the impedance characteristics. Last but not least, the modified admittance controller implements the motion of the robot using the data provided by the other parts. The framework exhibits a reliable robustness to tiny offset of position, making a wider range of applications possible.

However, the current framework has also potential to be improved further. The proper number of BFs depends highly on the length of demonstrated trajectory i.e. the duration of the demonstration. Thus, the autonomous adaption of number of BFs could be considered in the future. Besides, the reward function of RL could be improved to reach a better performance according to different scenarios. Also the hyperparameters of the RL and DMPs could be further researched to make the training faster and better.


\linespread{1.11}
\balance
\bibliographystyle{IEEEtran}
\bibliography{lib, bibliography} 

\end{document}